\documentclass[DIV=12, 11pt, a4paper]{scrartcl}
\usepackage[T1]{fontenc}
\usepackage[numbers]{natbib}
\usepackage{tikz}
\usepackage{amsmath}
\usepackage{mathptmx}
\usepackage{amssymb}
\usepackage{scalerel}
\usepackage{bm}
\usepackage{microtype}
\usepackage{float}
\usepackage{booktabs}
\usepackage{url}
\usepackage{comment}
\usepackage{hyperref}

\newcommand{\autocite}[1]{\cite{#1}}

\usetikzlibrary{svg.path}

\definecolor{orcidlogocol}{HTML}{A6CE39}
\tikzset{
	orcidlogo/.pic={
		\fill[orcidlogocol] svg{M256,128c0,70.7-57.3,128-128,128C57.3,256,0,198.7,0,128C0,57.3,57.3,0,128,0C198.7,0,256,57.3,256,128z};
		\fill[white] svg{M86.3,186.2H70.9V79.1h15.4v48.4V186.2z}
		svg{M108.9,79.1h41.6c39.6,0,57,28.3,57,53.6c0,27.5-21.5,53.6-56.8,53.6h-41.8V79.1z M124.3,172.4h24.5c34.9,0,42.9-26.5,42.9-39.7c0-21.5-13.7-39.7-43.7-39.7h-23.7V172.4z}
		svg{M88.7,56.8c0,5.5-4.5,10.1-10.1,10.1c-5.6,0-10.1-4.6-10.1-10.1c0-5.6,4.5-10.1,10.1-10.1C84.2,46.7,88.7,51.3,88.7,56.8z};
	}
}

\newcommand\orcidicon[1]{\href{https://orcid.org/#1}{\mbox{\scalerel*{
				\begin{tikzpicture}[yscale=-1,transform shape]
				\pic{orcidlogo};
				\end{tikzpicture}
			}{|}}}}

\addtokomafont{disposition}{\rmfamily}

\begin{document}
	
\title{Estimating a Null Model of Scientific Image Reuse to Support Research Integrity Investigations}

\author{Daniel E. Acuna \orcidicon{0000-0002-7765-1595}\,, Ziyue Xiang \orcidicon{0000-0001-6054-5801}}

\date{}

\maketitle

\begin{abstract}
\noindent When there is a suspicious figure reuse case in science, research integrity investigators often find it difficult to rebut authors claiming that ``it happened by chance''. In other words, when there is a ``collision'' of image features, it is difficult to justify whether it appears rarely or not. In this article, we provide a method to predict the rarity of an image feature by statistically estimating the chance of it randomly occurring across all scientific imagery. Our method is based on high-dimensional density estimation of ORB features using 7+ million images in the PubMed Open Access Subset dataset. We show that this method can lead to meaningful feedback during research integrity investigations by providing a null hypothesis for scientific image reuse and thus a $p$-value during deliberations. We apply the model to a sample of increasingly complex imagery and confirm that it produces decreasingly smaller $p$-values as expected. We discuss applications to research integrity investigations as well as future work.

\vspace*{1em}
\end{abstract}

\section{Introduction}

Figure element reuse (also known as copy-move forgery) is a significant problem in science. Bik et al. \cite{bik2016prevalence} manually inspected a total of 20,621 biomedical research papers, estimating that about 3.8 percent of them may contain problematic figures. They also concluded that the prevalence of papers with problematic images has risen markedly during the past decade. In order to assist editors and research integrity investigators with uncovering suspicious reuses, Acuna et al. \cite{Acuna269415} proposed an automated reuse detection tool dedicated to this scenario. However, the fact that the software uncovers potential pairs of reuse does not indicate research integrity investigators can accuse authors of misconduct confidently. Moreover, when suspicious authors claim that the reuse happens ``at random'', it is usually very difficult to prove otherwise. As a result, for example, NSF has had many cases involving image fabrication, but few cases have resulted in misconduct verdicts \autocite{parrish2009image}. We clearly need additional tools to refute claims of copy by chance. While statistical tools to refute these claims, such as null hypothesis testing, are common in court during deliberations, they are not common in scientific image reuse investigations. 

\begin{figure}[htpb]
	\centering
	
	\includegraphics[width=0.6\linewidth, page=1]{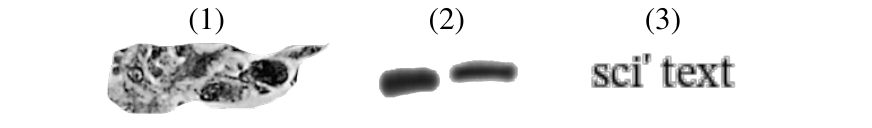}
	\caption{Assume these three image patches are suspicious reuse regions found by some software. These three patches contain microscopy, western blot and text, respectively. Intuitively, the suspicion decreases from left to right. But it is difficulty to justify it. The goal of this article is to provide an estimate of how likely a reuse can happen by chance by studying the probability distribution of image features. The solution to this problem is shown in Figure \ref{fig::targets-sol}.}
	\label{fig::targets}
\end{figure}

To illustrate the problem encountered while deliberating research integrity investigations, consider the situation illustrated by Figure \ref{fig::targets}. Imagine that these three image patches are found by reuse detection software. Panel 1, 2, and 3 correspond to a microscopy image, a western blot, and text, respectively. In other words, we assume that these image patches collide with some existing content in scientific publications. In the following text, ``reuse'' and ``collide'' are used interchangeably. Intuitively, the probabilities of such collisions occurring at random decreases from right to left. If the collided content is text, then it is likely to be a false alarm. If the collided content contains western blots, because blots look very similar, maybe it is a coincidence. If the suspicion cannot be eliminated, one can require raw data or ask the author to repeat the experiment. Nevertheless, if the collided patch consists of microscopy images (or other intricate content), then the probability of this happening at random is almost zero. These intuitions might inspire a technique to help during research integrity deliberations.

\begin{figure}[htpb]
	\centering
	\includegraphics[width=0.9\linewidth, page=2]{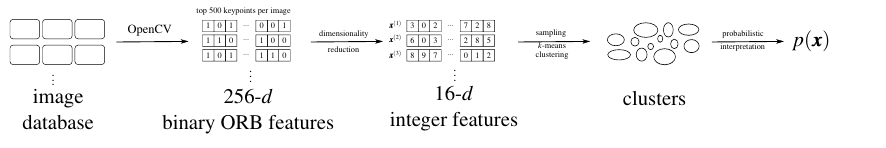}
	\caption{The flowchart of our work. We acquire all images within the PMC Open Access Subset and compute the top 500 ORB features \autocite{rublee2011orb} of each image. We apply dimensionality reduction to decrease the sparsity of ORB data, and the new feature vectors are randomly sampled without replacement. The resulting \textasciitilde $1.35\times 10^{8}$ records are fed into $k$-means clustering with 20k clusters. The distribution of ORB features is realized by employing probabilistic interpretation on the clustering results.}
	\label{fig::flowchart}
\end{figure}

The goal of this article is to determine the relationship between content and level of suspicion. We quantify the level of suspicion by measuring the probabilities of features within an image. The overall probabilities of an image patch determines how unlikely a reuse can happen at random, which allows research integrity investigators to reach a verdict with statistical confidence. The flowchart of our work is shown in Figure \ref{fig::flowchart}. We acquire all images in the PMC Open Access Subset\footnote{\url{https://www.ncbi.nlm.nih.gov/pmc/tools/openftlist/}} and compute the top 500 ORB features \autocite{rublee2011orb} for each image. We apply dimensionality reduction techniques to reduce the sparsity of ORB features, which also transforms the binary ORB features into continuous ones. In order to estimate the probability density of all ORB features, we run a large-scale $k$-means algorithm on the sampled dataset and employ probabilistic interpretation on the clustering.

In section \ref{sec::orb}, we briefly introduce the mechanisms of ORB features. In section \ref{section::model_dist}, we discuss how to apply dimensionality reduction to ORB features and how to build a probability distribution from the $k$-means clustering. In section \ref{sec::comp_vis}, we discuss details of our implementation and analyze the results of our experiments. In section \ref{sec::conclusion}, we provide a conclusion for this work.

\section{Describing Image Keypoints with ORB Features}\label{sec::orb}

A key component of our method is to compute mathematical descriptions of image features. Ideally, one image should have multiple such descriptors because there could be multiple areas of the images being reused. In this section, we will describe how we use ORB, a type of image keypoint descriptor, for this task.

ORB is an efficient, public-domain image keypoint detector and descriptor \autocite{rublee2011orb}. Compared to other methods such as SIFT \autocite{lowe2004distinctive} and SURF \autocite{bay2006surf}, ORB achieves similar performance with smaller time consumption \autocite{karami2017image}. The two building blocks of ORB are the FAST keypoint detector \autocite{rosten2006machine, rosten2008faster} and the BRIEF feature descriptor \autocite{calonder2010brief}. We now describe both.

As suggested by its name, FAST detectors are widely used because of the rapid computation. However, it does not provide an orientation component. ORB suggests using the intensity centroid \autocite{rosin1999measuring} as an inexpensive way to estimate the angle of a keypoint. The moments of a patch is given by
\begin{align*}
m_{pq} = \sum_{x, y} x^p y^q I(x, y),
\end{align*}
where $I(\cdot)$ denotes the intensity of the pixel at a given point. The intensity centroid of the patch is defined as
\begin{align*}
P = \left(\frac{m_{10}}{m_{00}}, \frac{m_{01}}{m_{00}}\right).
\end{align*}
We can construct a vector from the corner of the patch (denoted by $O$) to $P$, where the orientation of the patch is given by the orientation of $\overrightarrow{OP}$. It can be shown that the orientation $\theta$ is
\begin{align*}
\theta = \operatorname{atan2}\left(\frac{m_{10}}{m_{00}}, \frac{m_{01}}{m_{00}}\right) = \operatorname{atan2}(m_{01}, m_{10}),
\end{align*}
where $\operatorname{atan2}$ is the quadrant-aware $\arctan$ function. Since FAST does not produce multi-scale features, a scale image pyramid is built where FAST keypoints are detected at each level. Also, due to the fact that FAST does not produce a measure of ``cornerness'', the Harris corner measure \autocite{rosin1999measuring} is introduced to rank the quality of each detected keypoint.

After acquiring the keypoints, their characteristics are computed by the BRIEF descriptor. The BRIEF descriptor is a bit string representation of an image constructed from a predefined set of binary intensity tests. More concretely, assume there are $T$ point pairs $(\bm{z}^{(1)}_1, \bm{z}^{(1)}_2), \ldots, (\bm{z}^{(T)}_1, \bm{z}^{(T)}_2)$, where $\bm{z}^{(i)}_j = (x^{(i)}_j, y^{(i)}_j)$, the BRIEF descriptor is a $T$ dimensional binary vector $\bm{B}$. The $i$-th component of $\bm{B}$, which is denoted by $\bm{B}_i$, is given by
\begin{align*}
\bm{B}_i = 
\begin{cases}
1, &\mbox{if}~I(\bm{z}^{(i)}_1) < I(\bm{z}^{(i)}_2)\\
0, &\mbox{if}~I(\bm{z}^{(i)}_1) \geq I(\bm{z}^{(i)}_2).
\end{cases}
\end{align*}
Although BRIEF features are easy and fast to compute, they are not robust against rotations. Calonder et al. \cite{calonder2010brief} point out that this problem can be solved by creating a set of rotations and perspective wraps of the image patch, which is computationally expensive. An alternative solution, which is called \emph{steered} BRIEF, transforms the predefined point pairs instead of the patch. In this case, we only need to apply all transformations on the point pairs once to build up a series of precomputed BRIEF templates and then choose the best template during the actual computation step, eliminating the need of recalculating the image. Eventually, the BRIEF template which has the rotation angle closest to $\theta$ will be used to compute the BRIEF descriptor.

We compute ORB features with the OpenCV library \autocite{opencv_library}, where the value of $T$ is set to be 256 by default.

\section{Modeling the Distribution of ORB Features}\label{section::model_dist}

Although for our purpose it is intuitive to use a generative clustering model such as Gaussian Mixture Model (GMM) on the ORB features, the large amount of data and clusters make it unfeasible to solve the corresponding optimization problem. We make our solution scalable by reducing the dimensionality of ORB features and approximating the mixture model with a fast $k$-means algorithm and a conditional independence assumption of the joint feature distribution. We now describe these steps in detail.

\subsection{Dimensionality reduction}

We observe that the ORB feature space is sparse, as it is supposed to be so for effective feature matching. In order to study its distribution, we can decrease its sparsity by reducing the dimensionality. We apply dimensionality reduction by grouping every 16 bits of the ORB feature into one integer that counts the number of ones in the bits. In this way, the 256 bit binary ORB feature generated by OpenCV is turned into a 16 dimensional feature with numeric values, which allows the use of $k$-means algorithm. 

Since feature matching with ORB is done by comparing the hamming distance, which is essentially the squared Euclidean distance between bits, it would be desirable that the hamming distance of two ORB feature vectors before transformation is strongly linearly correlated with the squared Euclidean distance after transformation. To study the correlation, for each hamming distance $d=1, 2, \ldots, 30$, we generate 200,000 pairs of random ORB feature vectors that have hamming distance $d$, and then apply dimensionality reduction to them. Then, we compute the Pearson correlation coefficient between the hamming distance of original vectors and the squared Euclidean distance of transformed vectors. The value of correlation coefficient acquired is $\rho = 0.807$, $p < 0.0001$, which indicates a strong linear correlation between the two quantities. This suggests that the dimensionality reduction process preserves the similarity of ORB features to a great extent.

In the subsequent discussion, the terms ``sample'' and ``feature vector'' both refer to the ORB features after dimensionality reduction.

\subsection{Probabilistic interpretation of $k$-means clustering}

\begin{table}[htpb]
	\centering
	\begin{tabular}{cl}
		\toprule
		Symbol & \multicolumn{1}{c}{Description}\\ \midrule
		$\bm{x}^{(t)}_{s}$ & the $s$-th component of $t$-th sample\\
		$K$ & number of clusters of $k$-means clustering\\
		$C_j$ & the $j$-th cluster\\
		$\lvert C_j \rvert$ & the size of $j$-th cluster\\
		$N$ & total number of samples\\ 
		$M$ & the size of a feature vector\\
		\bottomrule
	\end{tabular}
	\vspace*{1em}
	\caption{List of symbols used in this subsection}
\end{table}

We now describe how we turn our density estimation problem to one of clustering and probability interpretation. We first apply a $k$-means algorithm to the dimensionality-reduced ORB features. Assume that we set the number of clusters in the $k$-means algorithm to be $K$. Denote the $j$-th cluster by $C_j$, $j = 1, \ldots, K$. $k$-means will assign a certain number of samples to each cluster, that is
\begin{align*}
C_j &= \varnothing, \quad \mbox{or}\\
C_j &= \{\bm{x}^{(j_1)}, \ldots, \bm{x}^{(j_{\lvert C_j \rvert})}\},
\end{align*}
where $\lvert C_j \rvert$ denotes the size of $C_j$. Let $N$ be the total number of samples, it is easy to see that
\begin{align*}
N = \sum_{j = 1}^{K} \lvert C_j \rvert.
\end{align*}
For each cluster $j$, we would like to model $p(\bm{x} \mid C_i)$. It is common to apply the conditional independence assumption, which indicates
\begin{align*}
p(\bm{x} \mid C_j) = \prod_{i = 1}^{M} p(\bm{x}_i \mid C_j),
\end{align*}
where $\bm{x}_i$ denotes the $i$-th component of the feature vector and $M$ denotes size of the feature vector. Since each $\bm{x}_i$ takes on discrete values, we can model $p(\bm{x}_i \mid C_j)$ with categorical distribution. To cope with missing values, we apply add-one smoothing to the parameter estimations. Therefore, we have
\begin{align*}
p(\bm{x}_i = a\mid C_j) = \frac{\left(\sum_{s = 1}^{\lvert C_j \rvert} \bm{1}\{ \bm{x}^{(j_s)}_i = a \} \right) + 1}{\lvert C_j \rvert + l},
\end{align*}
where $\bm{1}\{\cdot\}$ is the indicator function. Once we can compute $p(\bm{x} \mid C_j)$, by the law of total probability, we can write
\begin{align*}
p(\bm{x}) = \sum_{j = 1}^{K} p(\bm{x} \mid C_j)p(C_j), 
\end{align*}
where
\begin{align*}
p(c_j) = \frac{\lvert C_j \rvert}{N}.
\end{align*}
In order to estimate the scale of $p(\bm{x})$ more precisely with floating point arithmetic, we use the value of $\ln p(\bm{x})$ instead of $p(\bm{x})$. Meanwhile, the log-sum-exp trick is applied as follows:
\begin{align*}
p(\bm{x}) &= \sum_{j = 1}^{K} \exp\left\{\ln\left[p(\bm{x} \mid C_j)p(C_j)\right]\right\}\\
&= \sum_{j = 1}^{K} \exp\left\{\ln p(\bm{x} \mid C_j) + \ln p(C_j)\right\}\\
&= \sum_{j = 1}^{K} \exp\left\{\sum_{i=1}^{M}\ln p(\bm{x}_i \mid C_j) + \ln p(C_j)\right\}.
\end{align*}
Therefore, if we let $h_j(\bm{x})$ be
\begin{align*}
h_j(\bm{x}) = \sum_{i=1}^{M}\ln p(\bm{x}_i \mid C_j) + \ln p(C_j),
\end{align*}
a more numerically stable value of $\ln p(\bm{x})$ is given by
\begin{align*}
\ln p(\bm{x}) &= \ln \left[\sum_{j = 1}^{K} e^{h_j(\bm{x})}\right]\\
&= \operatorname{LogSumExp}\left(h_1(\bm{x}), \ldots, h_K(\bm{x})\right).
\end{align*}

\section{Results}\label{sec::comp_vis}

\subsection{Getting the distribution}
We acquire all 7,636,156 figures inside the PMC Open Access Subset (summer 2019 copy), which are considered the sample space of all images. Then, we compute ORB features for the top 500 keypoints within each image and apply the dimensionality reduction technique introduced in section \ref{section::model_dist}. For the $k$-means algorithm, we use the \texttt{kmcuda} software \autocite{vadim_markovtsev_2017_286944}, which supports multiple GPU acceleration and a large number of clusters. In our experiment, $K$ is set to 20,000. The seeds of $k$-means are selected with \texttt{k-means++} to find better clusterings \autocite{arthur2006k}. Due to memory constraints, the clustering algorithm runs on a randomly sampled subset of all data points, which contains $1/27$ of all records. The sampled subset contains approximately 135 million records.

\begin{figure}[htpb]
	\centering
	\includegraphics[width=0.6\linewidth, page=3]{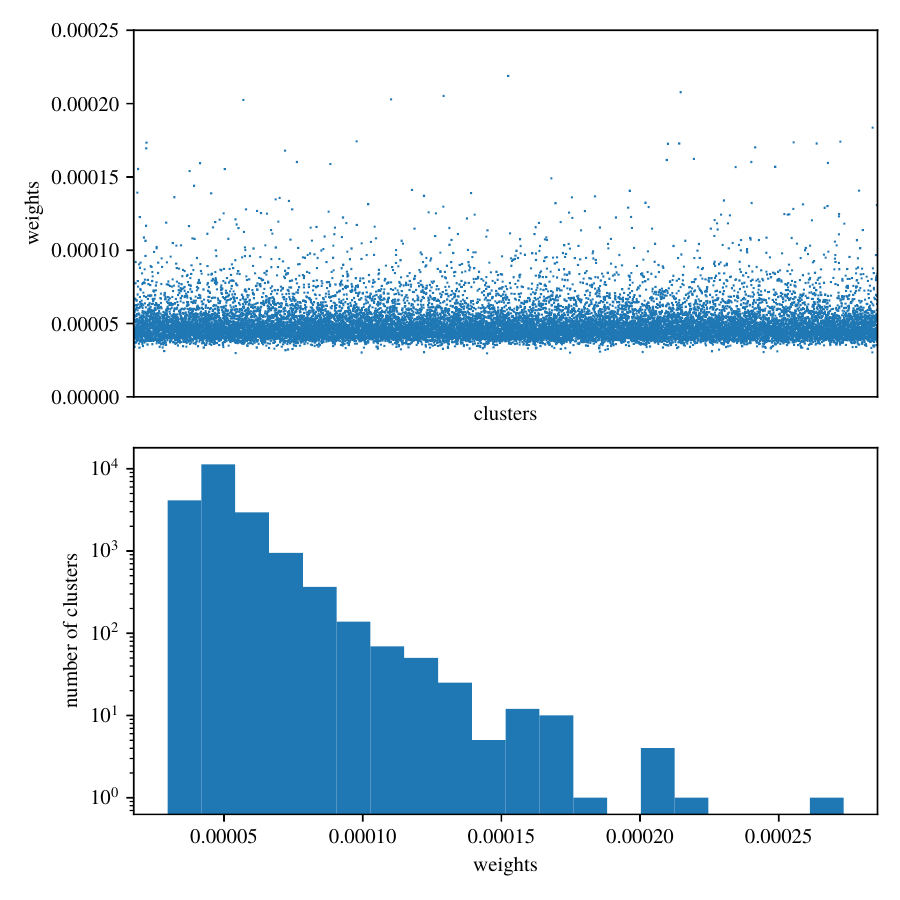}
	\caption{The weights of each cluster and histogram of weights after $k$-means clustering.}
	\label{fig::cluster_weights}
\end{figure}

After clustering, the weight of each cluster $\left(p(C_j)\right)$ and their histogram are shown in Figure \ref{fig::cluster_weights}. It can be seen that the weights of most clusters are between 0.00004 and 0.00006, while there are are number of clusters whose weights are greater than this range.

\subsection{Computing the ORB feature of any point in an image}
So far, we have tried to acquire the distribution of top 500 keypoints within a known image in the data base. Given a new image, we are interested in evaluating the similarity between every single point in the image and existing keypoints in the database, which is described in terms of probabilities. 

To compute the ORB feature of an arbitrary point in the image, we can apply modifications to OpenCV's ORB feature computation process. More specifically, for a given point in an image, we compute its Harris response in each layer of the image pyramid and select the layer with the highest Harris response. The orientation and BRIEF descriptor are then computed within the best layer. 

Since it is computationally expensive to estimate the probability of every single point in the image, we sample points from the image uniformly and then apply linear interpolation to the results. In our implementation, the sample distance is 3 pixels.

\subsection{Analyzing the outcome}

\begin{figure}[htpb]
	\centering
	\includegraphics[width=0.95\linewidth, page=6]{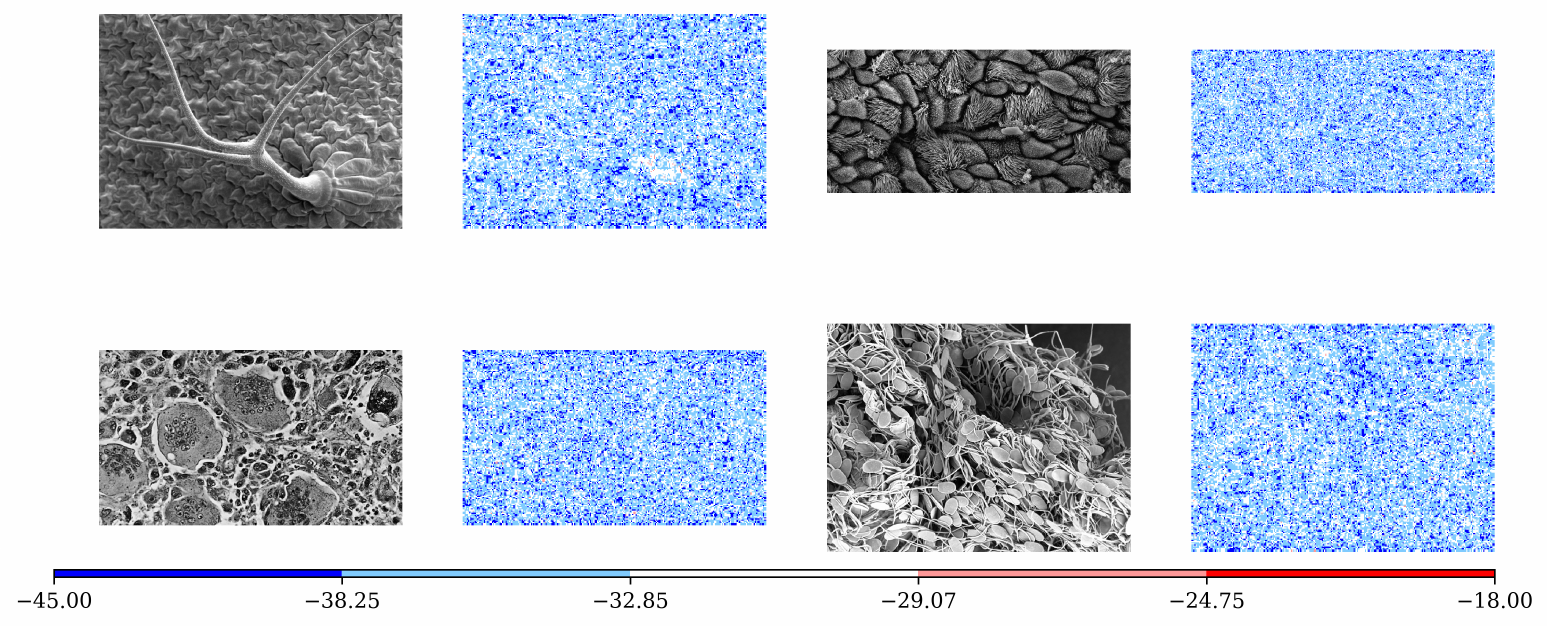}
	\caption{Examples showing the log probability of microscopy images.}
	\label{fig::mc_ex}
\end{figure}

\begin{figure}[htpb]
	\centering
	\includegraphics[width=0.95\linewidth, page=7]{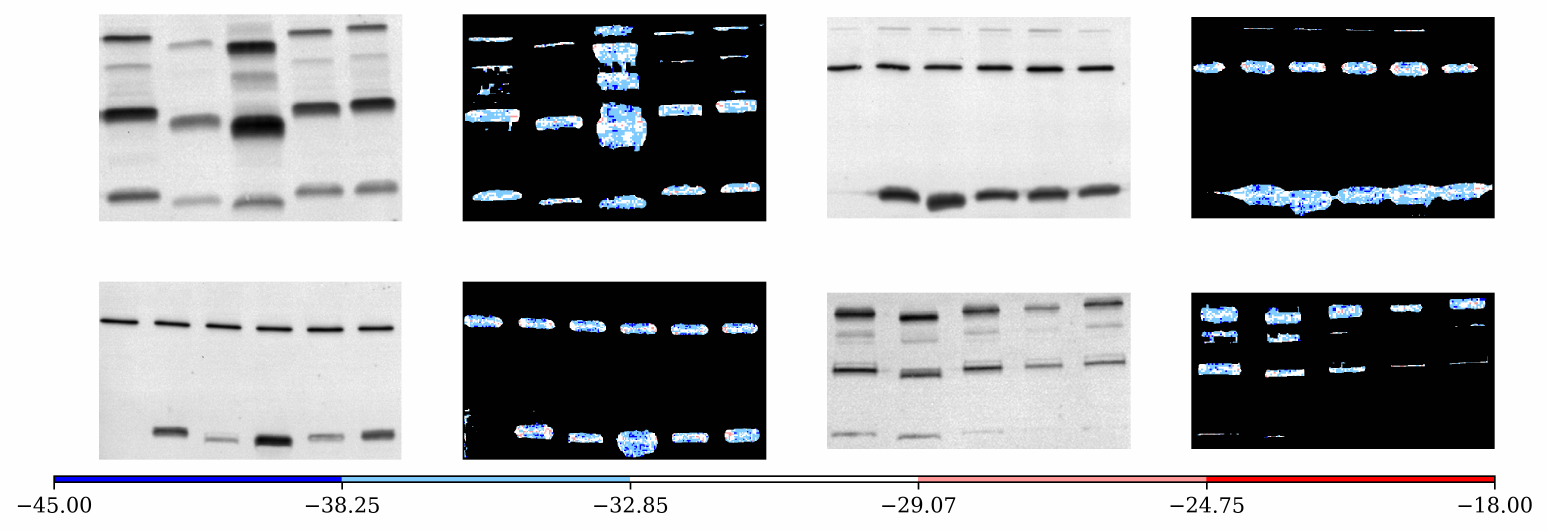}
	\caption{Examples showing the log probability of western blot images. Irrelevant regions are masked by black color.}
	\label{fig::blot_ex}
\end{figure}

\begin{figure}[htpb]
	\centering
	\includegraphics[width=0.95\linewidth, page=8]{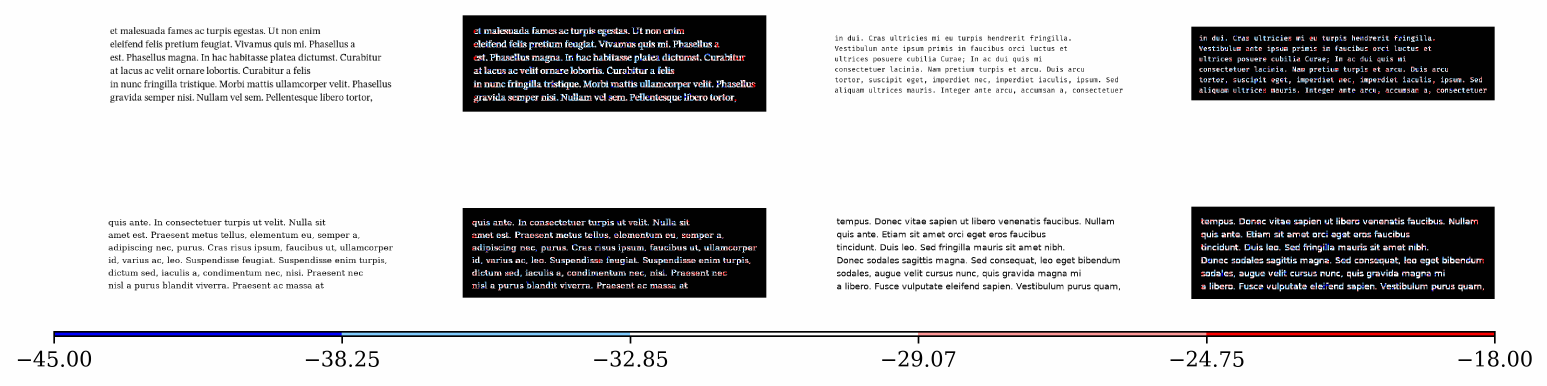}
	\caption{Examples showing the log probability of text images. Irrelevant regions are masked by black color.}
	\label{fig::text_ex}
\end{figure}

To verify the validity of our method, we select images that only contain three prevalent figure elements in scientific publications and compute the probability of image features inside them. These three types of elements are microscopy (which is close to natural images), western blot and text. Examples of microscopy, western blot and text are shown in Figure \ref{fig::mc_ex}, \ref{fig::blot_ex} and \ref{fig::text_ex} respectively. In Figure \ref{fig::mc_ex} and \ref{fig::blot_ex}, irrelevant regions are masked by black color.

\begin{figure}[htpb]
	\centering
	\includegraphics[width=0.6\linewidth, page=4]{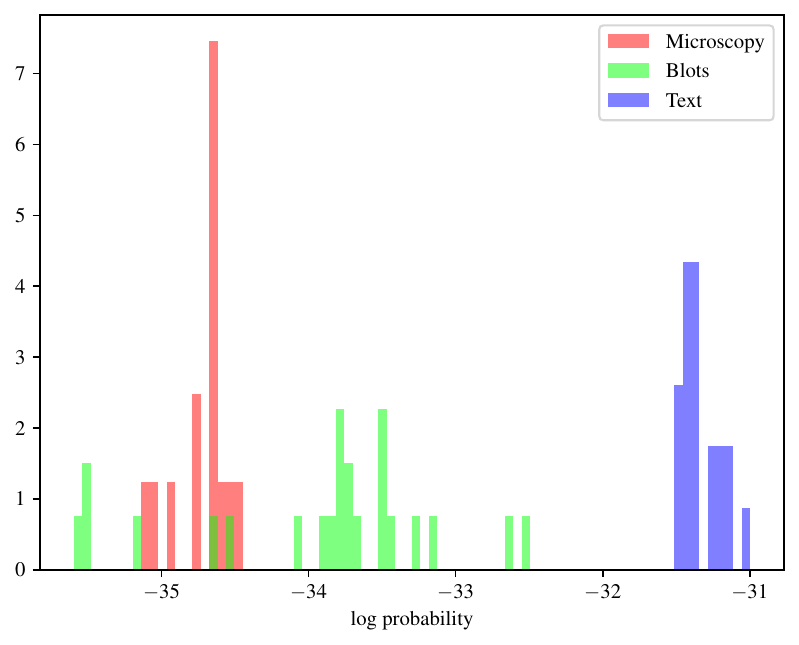}
	\caption{The histogram of image feature probabilities across three types of images.}
	\label{fig::hist}
\end{figure}

From the results, we can see that microscopy photos are rich in low probability features, for dark blue points can be seen in the probability map frequently. For western blots, their image features are of higher probability, as light blue and white are the dominant colors in the probability map. Features from text images have the highest probability, because red points appear regularly. This is reasonable considering the nature and the frequency of these three types of features in scientific images. As we plot the histogram of image feature probabilities (Figure \ref{fig::hist}), it can be seen that these three groups of features possess three distinctive distributions. To further study the significance of the result, we compute the mean log probability of each image and use them to form observations for each group. Then, a series of Welch's $t$-tests (Table \ref{table::t_test}) are conducted to measure the difference among the distributions, which generate more robust results compared to standard $t$-tests \autocite{derrick2016welch}. The $p$-values imply with strong confidence that the mean probabilities of image features inside distinctive image groups are different. This conform with our intuitions, as the contents of microscopy images are highly complicated and unlikely to be duplicated within the dataset, which accounts for their lower probability. For blots and text, however, they should have greater probabilities due to higher intra-group feature similarity.

\begin{table}[htpb]
	\centering
	\begin{tabular}{cc}
		\toprule
		Test objects & $p$-value\\ \midrule
		microscopy \& western blots & $3.045\times10^{-4}$\\
		microscopy \& text & $7.825\times10^{-26}$\\
		text \& western blots & $3.422\times10^{-13}$\\
		\bottomrule\\
	\end{tabular}
	\caption{$p$-values of Welch's t-test among the average log probability of each image group.}
	\label{table::t_test}
\end{table}

\begin{figure}[htpb]
	\centering
	\includegraphics[width=0.6\linewidth, page=5]{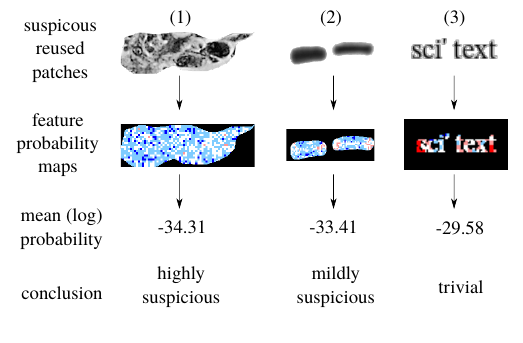}
	\caption{Examples of using the distribution of ORB features to support figure element reuse investigation. }
	\label{fig::targets-sol}
\end{figure}

Back to the problem raised by Figure \ref{fig::targets}. Our proposed solution to it is shown in Figure \ref{fig::targets-sol}. After computing the probability map of image features within a given patch, a research integrity investigator can reach a fact-supported verdict by visually inspecting the output map or by analyzing the mean log probabilities.

\section{Discussion}\label{sec::conclusion}

In this paper, we proposed a tool that gives probabilities of image features to support figure element reuse investigations. The probability distribution is acquired by employing probabilistic interpretation on the $k$-means clustering of ORB features acquired from the PMC Open Access Subset. To reduce the sparsity of ORB features for better clusterings, we apply similarity-preserving dimensionality reduction to ORB features. The statistical tests on the results of experiments show that various groups of image contents tend to have distinctive probability distributions, and that estimated probabilities of these groups conform with our intuitions. In conclusion, our method is able to accurately report a level of suspicion given reused image content.

Due to computer hardware constraints, we are unable to analyze the entire dataset. In the dimensionality reduction step, we choose to group 16 bits into one integer as an attempt to prepare the data for $k$-means clustering and reduce the size of ORB feature vectors. When less bits are aggregated together, the distance will be preserved better after the transformation, but it will also increase the vector size and reduce the number of records in the $k$-means step. Generally, two types of errors are introduced in our analysis procedure: the first one is due to the inability to view the entire dataset; the second one is due to the skewed distance between transformed ORB vectors. Unfortunately, given limited computational resource, it is impossible to eliminate both errors completely, for the decrease in one error will eventually result in the increase of another. This prevents us from acquiring a more precise result. In addition, the output of our method is the log probability of an image feature, which does not reflect the probability of reuse directly. Additional interpretation needs to done to determine the relationship between image feature probability and reuse probability.

\subsection{Best practices}

Our work provides a useful tool for research integrity investigation, as now findings can be associated with a degree of confidence. It can also serve as a filter for copy-move forgery detection software so that trivial matches can be discarded automatically. When there is a figure reuse accusation, our method can improve the quality of evidence and also help protect the defendant using a data driven approach. We hope that our method can lead to more data-driven investigations and therefore help both the accused and research investigators. Future work will adapt our framework to other modes such as text, equations, statistical analysis, and citations.

\subsection{Research Agenda}

In our opinion, improvements can be made in the following aspect, which are able to generate more accurate results.
\begin{itemize}
	\item Data source: this study focuses on the PMC Open Access dataset, which is a small subset of all scientific images. A more comprehensive distribution can be acquired if more datasets are considered.
	\item Image feature: we only use a specific type of image feature (ORB). It would be interesting to study the distribution of SIFT, SURF and some novel deep-learning based image features \cite{yang2018multi, ono2018lf}.
	\item Hardware: the hardware performance limits our ability to process all data with high accuracy. With better hardware, one is likely to acquire more precise output.
\end{itemize}

\subsection{Educational implications}

The digital age makes information more accessible than ever before, which lowers the bar and detectability of research misconduct. It is becoming increasingly difficult for the science community to monitor research integrity using traditional approaches, for the amount of knowledge involved simply exceed humans' capacity. However, with the rapid development of AI and big data analytics, there may be a way for one to pinpoint the evidence inside terabytes of potentially relevant data. We propose and advance the concept of ``computational research integrity'', as it provides new and reliable tools for research integrity investigators. We also hope that more researchers can participate in this emerging and promising field.

\subsection{Acknowledgments}

Daniel E. Acuna and Ziyue Xiang are funded by the Office of Research Integrity grants \#ORIIR180041 and \#ORIIR19001.

\bibliography{ref}

\end{document}